\documentclass[letterpaper, conference, twoside]{IEEEtran}
\usepackage[utf8]{inputenc}
\IEEEoverridecommandlockouts

\usepackage{amsmath}
\usepackage{amssymb}
\usepackage{amsfonts}

\usepackage{booktabs}
\usepackage{subcaption}
\usepackage{hyperref}
\usepackage[dvipsnames]{xcolor}

\usepackage{graphicx}
\graphicspath{img/}

\usepackage{tikz}
\usetikzlibrary{positioning}
\usepackage{flushend}
\usepackage{cite}

\newcommand{\vect}[1]{\mathbf{ #1}}

\newcommand{\vectg}[1]{{\boldsymbol{ #1}}}

\newcommand{\R}{\mathbb{R}}

\newcommand{\T}{^\mathsf{T}}

\newcommand{\argmin}{\operatornamewithlimits{argmin}}

\newcommand{\vQ}{\vect{Q}}
\newcommand{\vR}{\vect{R}}

\newcommand{\vZ}{\vect{Z}}

\newcommand{\vb}{\vect{b}}

\newcommand{\vm}{\vect{m}}

\newcommand{\vq}{\vect{q}}

\newcommand{\vs}{\vect{s}}
\newcommand{\vt}{\vect{t}}
\newcommand{\vu}{\vect{u}}

\newcommand{\vw}{\vect{w}}
\newcommand{\vx}{\vect{x}}

\newcommand{\vz}{\vect{z}}

\newcommand{\vbeta}{\vectg{\beta}}

\newcommand{\vpi}{\vectg{\pi}}

\newcommand{\veta}{\vectg{\eta}}

\newcommand{\vtheta}{\vectg{\theta}}

\newcommand{\cN}{\mathcal{N}}

\newcommand{\etal}{\textit{et al}.\ }
\newcommand{\ie}{\textit{i}.\textit{e}.\ }
\newcommand{\eg}{\textit{e}.\textit{g}.\ }

\title{An Orientation Factor for Object-Oriented SLAM}

\author{Natalie Jablonsky, Michael Milford, and Niko S{\"u}nderhauf%
\thanks{The authors are with the ARC Centre of Excellence for Robotic Vision, Queensland University of Technology (QUT), Brisbane, Australia (e-mail: natalie.jablonsky@hdr.qut.edu.au; michael.milford@qut.edu.au; niko.suenderhauf@qut.edu.au).}
\thanks{This research was conducted by the Australian Research Council Centre of Excellence for Robotic Vision (project number CE140100016). Michael Milford is supported by an Australian Research Council Future Fellowship (FT140101229).}
}
\begin{document}

\maketitle

\begin{abstract}
    Current approaches to object-oriented SLAM lack the ability to incorporate prior knowledge of the scene geometry, such as the expected global orientation of objects.  
    We overcome this limitation by proposing a \emph{geometric factor} that constrains the global orientation of objects in the map, depending on the objects' semantics. This new geometric factor is a first example of how semantics can inform and improve geometry in object-oriented SLAM. We implement the geometric factor for the recently proposed QuadricSLAM that represents landmarks as dual quadrics. The factor probabilistically models the quadrics' major axes to be either perpendicular to or aligned with the direction of gravity, depending on their semantic class. 
    Our experiments on simulated and real-world datasets show that using the proposed factors to incorporate prior knowledge improves both the trajectory and landmark quality.
\end{abstract}

\begin{figure}[t]
    \centering
    \includegraphics[width=0.45\textwidth]{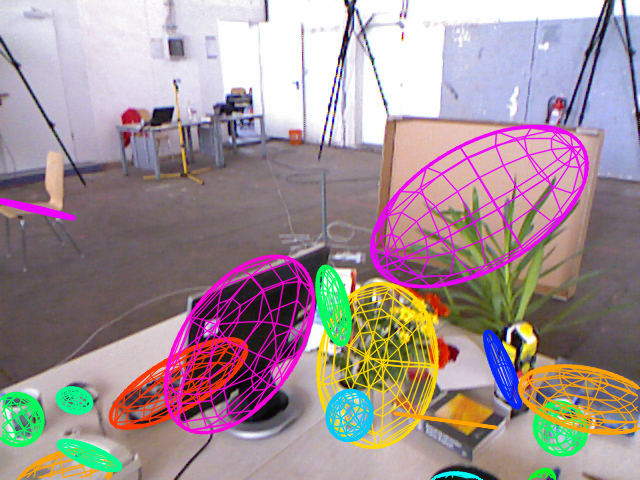}\vspace{0.2cm}\\%
    \includegraphics[width=0.45\textwidth]{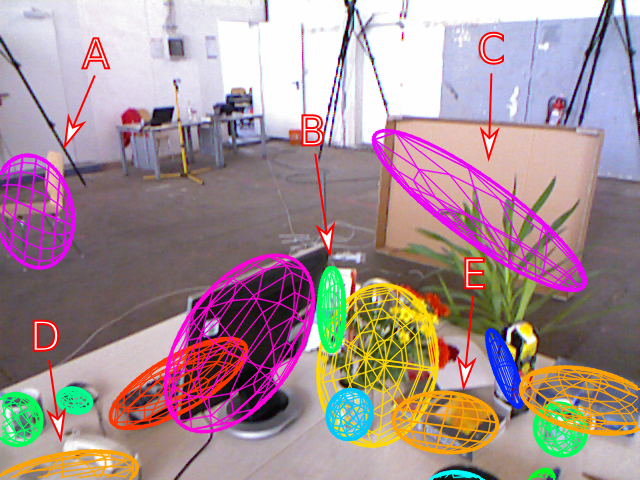}
    \caption{Comparing standalone QuadricSLAM~\cite{Nicholson2018} (\emph{top}) versus QuadricSLAM with both orientation factors proposed in this paper (\emph{bottom}). 
    In this example our new factors visibly improve the estimated quadrics around the chair (A), bottle (B), phone (D) and book (E), resulting in  a better alignment with the true objects. The cardboard box (C) was misclassified as a chair by the object detector (YOLO) which resulted in the wrong orientation factor being applied.}
    \label{fig:before_vs_after}
\end{figure}

\section{Introduction} \label{sec:intro}
Fuelled by the advances in visual object detection and semantic segmentation \cite{YOLO,SSD,Yolo9000,he2018mask}, research in robotic mapping and SLAM has recently focused on creating maps that not only capture the geometry of a scene, but also its semantics \cite{mccormac2017semanticfusion, McCormac2018,sunderhauf2017meaningful,Nicholson2018}. Semantically-enriched maps are appealing for many robotic applications, as they increase the capabilities of a robot to understand and reason about the complex world around it. This consequently increases the range and sophistication of interactions that robots can have with the  world, a critical requirement for their eventual widespread deployment at work and in homes.

QuadricSLAM \cite{Nicholson2018} is a recently proposed object-oriented SLAM system that represents objects as quadrics (spheres and ellipsoids) which are estimated from visual object detections. Quadrics provide a compact representation that approximates an object's position, orientation and shape. 
A shortcoming of QuadricSLAM is that it cannot incorporate prior knowledge about the expected global orientation of the mapped objects. This can result in inaccurate quadric estimates, as illustrated in Fig.~\ref{fig:before_vs_after}.
Another recent approach to object-oriented SLAM \cite{Hosseinzadeh2018} has proposed constraints that exploit geometric prior knowledge of the world's structure but lacks an integration of semantics into the SLAM system. The recent work of Atanasov \etal \cite{Atanasov2018} moves towards leveraging semantics to inform the geometry by constraining the structure of object parts that form the object as a whole. But their approach does not use the explicit semantic labels of object parts in their constraint.

Correctly estimating the position, orientation and shape of objects is critical if we want a robot to perform a comprehensive reasoning of a scene. For the majority of indoor scenes, objects are stable and are supported by planar surfaces. Meaningful maps must represent objects in physically plausible positions if a robot is going to interact with objects. 

In this work, we introduce a geometric factor that constrains objects' global orientation to be either perpendicular to or aligned with the direction of gravity. For QuadricSLAM this is done by using the cosine similarity between the quadric's major axis and the $z$-axis in the global frame. The vertical orientation aligns in the direction of the $z$-axis while the horizontal is perpendicular to the $z$-axis. We use object class labels from the object detector YOLO~\cite{YOLO} to determine the expected orientation of the objects, enabling semantics to inform the geometry. 

We demonstrate the performance of this orientation factor on simulated and real-world datasets, measuring the trajectory and landmark quality. We show that incorporating all orientations and exploiting prior knowledge improves the trajectory and landmark quality.

\section{Related Work} \label{sec:related_work}
Current SLAM systems are predominantly focused on capturing the geometric structure of the environment. Maps created by state-of-the-art systems like ORB-SLAM \cite{MurArtal15tro,Mur-Artal2016}, LSD-SLAM \cite{Engel2014} and ElasticFusion \cite{Whelan15rss} lack semantic information which could be leveraged to enable robots to complete more complex tasks. Recent advances in object-oriented and semantic mapping have made progress in addressing classification, detection and segmentation \cite{sunderhauf2017meaningful,mccormac2017semanticfusion,McCormac2018}. However these approaches do not take full advantage of the relationship between the geometry and the semantics of a scene. The recent development of \cite{Atanasov2018} has shown the feasibility of coupling semantics and geometry into SLAM.

Our paper builds on the work of Nicholson \etal \cite{Nicholson2018} which introduced an object-oriented SLAM system that estimates quadric landmarks from the outputs of a visual object detector (YOLO~\cite{YOLO}). 
The quality of the quadric estimates are negatively affected by high variance in bounding box measurements of YOLO, object occlusions and false object detections. 
In our work, we show that by 
imposing semantically-informed orientation constraints on quadrics we can improve both the landmark and trajectory quality. 

The related work in \cite{Hosseinzadeh2018} investigated the use of quadric landmarks in conjunction with points and planes. Their method proposed the use of these geometric entities imposed with structural constraints to exploit prior knowledge of the environment. 
However, their method also does not take advantage of semantic information to influence the geometry of the map. 

\section{QuadricSLAM} \label{sec:quadric_slam}
This section outlines the core concepts of dual quadrics as well as the general formulation of QuadricSLAM. These concepts will be used throughout the remainder of the paper. For the sake of completeness, we refer the reader to \cite{Nicholson2018} for an in depth discussion of QuadricSLAM and its formulation.

\subsection{Dual Quadrics} \label{subsec:dual_quadrics}
In 3D projective space, quadrics are surfaces that are represented by a symmetric matrix $\vQ \in \R^{4 \times 4}$. In its dual form, a quadric is defined by a set of tangential planes such that the planes envelope the quadric. This dual quadric $\vQ^*$ is defined so that all planes $\vpi$ fulfil the equation $\vpi\T\vQ^*\vpi = 0$, where $\vQ^* = adjugate(\vQ)$ or $\vQ^* = \vQ^{-1}$ if $\vQ$ is invertible.

In its general form, a dual quadric can represent closed surfaces (such as spheres and ellipsoids) and non-closed surfaces (such as paraboloids or hyperboloids). Only closed surfaces are meaningful for object landmarks as paraboloids and hyperboloids can generate surfaces that are infinite. Therefore for QuadricSLAM, dual quadric representations are constrained to ensure that the represented surface is either an ellipsoid or sphere. Thus dual quadrics are parameterize as, 

\begin{equation}
    \vQ^* = \vZ \> \breve{\vQ}^* \> \vZ\T
    \label{eq:quadric_from_e}
\end{equation}

\noindent where $\breve{\vQ}^*$ is an ellipsoid centred at the origin, and $\vZ$ is a homogeneous transformation that accounts for an arbitrary rotation and translation. These matrices are defined as,

\begin{equation}
    \vZ =
    \begin{pmatrix}
        \vR(\vtheta) & \vt \\
        \bf{0}\T_{3} & 1 \\
    \end{pmatrix} \:\:\text{and}\:\:
    \breve{\vQ}^* = 
    \begin{pmatrix}
        s_{1}^2 & 0 & 0 & 0 \\
        0 & s_{2}^2 & 0 & 0 \\
        0 & 0 & s_{3}^2 & 0 \\
        0 & 0 & 0 & -1 \\
    \end{pmatrix}
\end{equation}

\noindent where $\vt = (t_1, t_2, t_3)$ is the quadric centroid translation, $\vR(\vtheta)$ is a rotation matrix defined by the angles $\vtheta = (\theta_1, \theta_2, \theta_3)$, and $\vs = (s_1, s_2, s_3)$ is the shape of the quadric along the three semi-axes of the ellipsoid. 
A constrained dual quadric can be compactly represent with a 9-vector $\vq = (\theta_1, \theta_2, \theta_3, \:t_1, t_2, t_3, \:s_1, s_2, s_3)\T$ which is used to reconstruct the full dual quadric $\vQ^*$ as defined in (\ref{eq:quadric_from_e}).

\subsection{QuadricSLAM problem}
For the QuadricSLAM problem there are two measurements. The first measurement is the set of the odometry measurements $U = \{\vu_i\}$ between two successive poses $\vx_i$ and $\vx_{i+1}$, such that $\vx_{i+1} = f(\vx_i, \vu_i) + \vw_i$. Where $f$ is the motion model of the robot and $\vw_i$ are zero-mean Gaussian error terms with corresponding covariances $\Sigma_i$.
 
The second measurement is the set of bounding box detections $B = \{\vb_{ij}\}$. Where the variable $\vb_{ij}$ indicates a bounding box around object $j$ being observed from pose $\vx_i$.

In building the factor graph \cite{Kschischang01it}, the conditional probability distribution over all robot poses $X = \{\vx_i\}$, and landmarks $Q = \{\vq_j\}$, given the observations of odometry $U = \{\vu_i\}$, and bounding box detections $B = \{\vb_{ij}\}$ can be factored as,

\begin{equation}
   P(X,Q|U,B) \propto \underbrace{\prod_i P(\vect{x}_{i+1} | \vect{x}_i,
   \vect{u}_{i})}_\text{Odometry Factors}
   \cdot
   \underbrace{\prod_{ij} P(\vq_j | \vect{x}_i, \vect{b}_{ij})}_\text{Landmark Factors}
  \label{eq:slam_posegraph_probability}
\end{equation}

With the set of observations $U$, $B$, the maximum a posteriori (MAP) of robot poses $X^*$ and dual quadrics $Q^*$ to solve the landmark SLAM problem is determined by maximizing the joint probability of \eqref{eq:slam_posegraph_probability}. This is formulated as a nonlinear least squares problem;

\begin{align}
  X^*, Q^* &=
      \argmin_{X,Q} -\log P(X,Q|U,B) \nonumber\\
      &=\argmin_{X,Q} 
      \underbrace{\sum_i \|f(\vect{x}_i, \vect{u}_i) \ominus \vect{x}_{i+1}\|^2_{\Sigma_{i}}}_\text{Odometry Factors}  \\
      & \hspace{1.2cm}+ \underbrace{\sum_{ij}  \|\vb_{ij} - \vbeta_{(\vx_i,\vq_j)}\|^2_{\Lambda_{ij}}}_\text{Quadric Landmark Factors} \nonumber
      \label{eq:SLAM_lsq}
\end{align}

\noindent where $\|a-b\|^2_\Sigma$ is the squared Mahalanobis distance with covariance $\Sigma$ and the $\ominus$ operator denotes the difference operation that is performed in SE(3).

\section{Geometric Factors} \label{sec:factors}
Humans intuitively know that some objects are typically oriented perpendicular (e.g. a bottle) or in parallel (e.g. a book) to their support surface. Robotic object-oriented SLAM systems like QuadricSLAM need to model such prior knowledge explicitly in a probabilistic way. In this section, we introduce a new geometric factor that captures prior knowledge about the expected global orientation of objects, depending on their semantic class.

The following global orientation factor makes use by imposing the major axis of quadric estimates to be vertical (aligning to the global $z$-axis) or horizontal (perpendicular to the $z$-axis).

\subsection{Finding the Major Semi-Axis of a Quadric}\label{subsec:major_axis}
The orientation of a quadric is translation invariant, therefore we are only concerned with the rotational and shape components of the quadric. Specifically we use the six parameters $\vq^{\prime} = (\theta_1, \theta_2, \theta_3, \:s_1, s_2, s_3)\T$ to reconstruct the quadric centred at the origin (\ie no translation). This is computed as,

\begin{equation} \label{eq:geometric_factor_quadric}
    \mathbf{Q^{*\prime}} = \vR(\vtheta) \begin{pmatrix}
        s^{2}_{1} & 0 & 0 \\
        0 & s^{2}_{2} & 0 \\
        0 & 0 & s^{2}_{3}
    \end{pmatrix}\vR(\vtheta)\T
\end{equation}

Using this representation of the quadric, the orientation of the quadric can be determined by finding the eigenvalues $\lambda = (\lambda_{1}, \lambda_{2}, \lambda_{3})$ and eigenvectors $\veta = (\veta_{1}, \veta_{2}, \veta_{3})$ of matrix $\mathbf{Q^{*\prime}}$. The eigenvector with the largest corresponding eigenvalue indicates the orientation of the major semi-axis of the quadric, which we will denote as $\vm$.

The bounded cosine similarity $c_{j}$ between the major semi-axis $\vm$ of quadric $j$ and the $z$-axis $\vz$ is used in the error of the quadric's orientation. Explicitly, this is defined as,

\begin{equation} \label{eq:cosine_sim}
    c_{j} = \Bigg| \frac{\vm_{j} \cdot \vz}{\|\vm_{j}\| \|\vz\|} \Bigg|
\end{equation}

\subsection{Vertical Orientation Error} \label{subsec:upright}
The vertical orientation aligns the major semi-axis of the quadric to the $z$-axis of the global coordinate frame. This is applied to objects whose major axis is typically aligned with the direction of gravity (\eg bottles, clocks and chairs). 

The error term for this orientation minimizes the cosine similarity between the major semi-axis $\vm$ and the global $z$-axis $\vz$. 

\begin{equation} \label{eq:upright_error}
    \| 1 - c_{j} \|^{2}_{\sigma}
\end{equation}

\subsection{Horizontal Orientation Error} \label{subsec:flat}
Similar to Section~\ref{subsec:upright}, we compute the eigenvalues and eigenvectors for the quadric to get the major semi-axis. The horizontal orientation then aligns the major semi-axis of a quadric to be perpendicular to the $z$-axis. This error function is defined as, 

\begin{equation} \label{eq:flat_error}
    \| 0 - c_{j} \|^{2}_{\sigma}    
\end{equation}

\subsection{Factor Graph Representation}
The joint probability of (\ref{eq:slam_posegraph_probability}) is updated to include the general form of the orientation factor. The orientation factor seeks to find the probability of the cosine similarity $c_{j}$ given the object's class label $l_{j}$. As such, the joint probability now takes the form,

\begin{align}
   P(X,Q|U,B) \propto &\underbrace{\prod_i P(\vect{x}_{i+1} | \vect{x}_i,
   \vect{u}_{i})}_\text{Odometry Factors}
   \cdot
   \underbrace{\prod_{ij} P(\vq_j | \vect{x}_i, \vect{b}_{ij})}_\text{Landmark Factors} \nonumber \\
   &\cdot
   \underbrace{\prod_{j} P(c_{j} | l_{j})}_\text{Orientation Factors}
  \label{eq:slam_posegraph_probability_updated}
\end{align}

The likelihood of the orientation factor is assumed to be Gaussian $\cN(g(l_{j}), \sigma)$, where the piecewise function $g(l_{j})$ determines the orientation of the quadric with respect to the class label,

\begin{equation}
    g(l_{j}) = \begin{cases} 
            0, & \text{if } l_{j} \in \text{horizontal} \\
            1, & \text{if } l_{j} \in \text{vertical}
        \end{cases}
\end{equation}

The nonlinear least squares problem of (\ref{eq:slam_posegraph_probability}) now includes the orientation factors. Formally, the optimal variable configuration of $X^{*}, Q^{*}$ that we want to solve is,

\begin{align}
  X^*, Q^* &= 
      \argmin_{X,Q} -\log P(X,Q|U,B) \nonumber\\
      &=\argmin_{X,Q} 
      \underbrace{\sum_i \|f(\vect{x}_i, \vect{u}_i) \ominus \vect{x}_{i+1}\|^2_{\Sigma_{i}}}_\text{Odometry Factors}  \nonumber\\
      &+ \underbrace{\sum_{ij}  \|\vb_{ij} - \vbeta_{(\vx_i,\vq_j)}\|^2_{\Lambda_{ij}}}_\text{Quadric Landmark Factors} \nonumber
      + \underbrace{\sum_{j}  \| g(l_{j}) - c_{j}  \|^2_{\sigma_{j}}}_\text{Orientation Factors} \\
      \label{eq:slam_lsq_updated}
\end{align}

\begin{figure}[t]
\centering
\begin{tikzpicture}[
var/.style={circle, draw=black!60, fill=black!5, very thick, minimum size=7mm},
odom/.style={rectangle, draw=RoyalBlue, fill=RoyalBlue, very thick, minimum size=2mm},
measure/.style={rectangle, draw=Dandelion, fill=Dandelion, very thick, minimum size=2mm},
horizontal/.style={rectangle, draw=Orange, fill=Orange, very thick, minimum size=2mm},
vertical/.style={rectangle, draw=Red, fill=Red, very thick, minimum size=2mm},
]
\node[measure, label=left:$b_{0,1}$]      (centre)                                        {};

\node[var]         (landmark1)     [above=of centre]               {$q_1$};
\node[var]         (landmark2)     [right=15mm of landmark1]       {$q_2$};
\node[var]         (landmark3)     [right=15mm of landmark2]       {$q_3$};

\node[measure, label=$b_{1,2}$]      (lf2)           [right=9mm of centre]          {};
\node[measure, label=right:$b_{1,3}$]      (lf3)           [below=of landmark2]            {};
\node[measure, label=left:$b_{2,4}$]      (lf5)           [below=of landmark3]            {};

\node[var]         (pose1)         [below=of centre]               {$x_0$};
\node[var]         (pose2)         [right=15mm of pose1]           {$x_1$};
\node[var]         (pose3)         [right=15mm of pose2]           {$x_2$};

\node[odom, label=$u_0$]      (pf0)           [left=6mm of pose1]             {};
\node[odom, label=$u_1$]      (pf1)           [right=6mm of pose1]            {};
\node[odom, label=$u_2$]      (pf2)           [right=6mm of pose2]            {};

\node[vertical, label=$c_{1}$]     (sf1)           [above=6mm of landmark1] {};
\node[vertical, label=$c_{2}$]     (sf2)           [above=6mm of landmark2] {};
\node[vertical, label=$c_{3}$]     (sf3)           [above=6mm of landmark3]       {};

\draw[-] (landmark1.south) -- (centre.north);
\draw[-] (centre.south) -- (pose1.north);
\draw[-] (landmark1.south) -- (lf2.north west);
\draw[-] (lf2.south east) -- (pose2.north);
\draw[-] (landmark2.south) -- (lf3.north);
\draw[-] (lf3.south) -- (pose2.north);
\draw[-] (landmark3.south) -- (lf5.north);
\draw[-] (lf5.south) -- (pose3.north);

\draw[-] (pf0.east) -- (pose1.west);
\draw[-] (pose1.east) -- (pf1.west);
\draw[-] (pf1.east) -- (pose2.west);
\draw[-] (pose2.east) -- (pf2.west);
\draw[-] (pf2.east) -- (pose3.west);

\draw[-] (landmark1.north) -- (sf1.south);
\draw[-] (landmark2.north) -- (sf2.south);
\draw[-] (landmark3.north) -- (sf3.south);
\end{tikzpicture}
\caption{The factor graph representation of QuadricSLAM with orientation factors. The factors included are the orientation factors $c_{j}$ (\emph{red}), odometry factors $u_{i}$ (\emph{blue}) and bounding box factors $b_{ij}$ (\emph{yellow}).}
\label{fig:factor_graph_rep}
\end{figure}
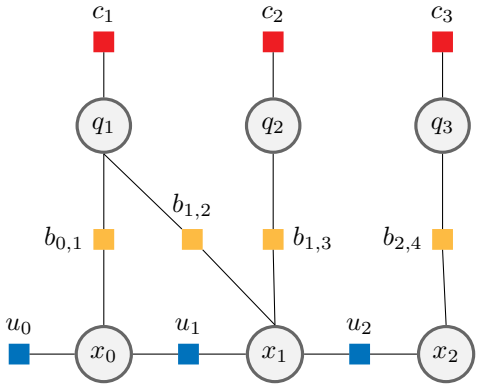

Figure~\ref{fig:factor_graph_rep} illustrates this least squares problem (\ref{eq:slam_lsq_updated}) as a factor graph.

\section{Experiments \& Evaluation} \label{sec:experiments}
We evaluated the use of orientation factors in QuadricSLAM on a number of sequences and datasets. The benchmark dataset TUM RGB-D \cite{Sturm2012a} was used to evaluate the estimated camera trajectory. Our own simulated environment was used to evaluate quadric landmark position, shape and quality against ground truth object positions. We compared our results with those of \cite{Nicholson2018} on both datasets. We also compared our TUM RGB-D results to ORB-SLAM2.

For both experiments we implemented QuadricSLAM and the orientation factors using GTSAM \cite{GTSAM}. The robot poses $X^*$ and dual quadrics $Q^*$ are the latent variables of the graph. Pose variables are connected with odometry factors $U$, while pose and dual quadric variables are connected with bounding box factors $B$. The orientation factors $C$ are connected to dual quadric variables based on their object classification.

Object classes used by the object detector were manually separated into horizontal, vertical or unassigned cases. Table~\ref{tab:classes_vs_factors} shows which orientation factor (if any) was applied to each object. These assumptions are based on prior knowledge of the sequences and how these objects are expected to be positioned within the environment.

\begin{table}[t]
    \centering
    \caption{Object classes and their assigned factors for the TUM sequences.}
    \label{tab:classes_vs_factors}
    \begin{tabular}{@{}ll|ll|l@{}}
        \toprule
        \multicolumn{2}{c}{Horizontal} & \multicolumn{2}{c}{Vertical} & Unassigned \\ \midrule
        Fork & Mouse & Person & Vase & Sports Ball \\
        Knife & Remote & Backpack & Potted Plant & \\ 
        Spoon & Keyboard& Bottle & Teddy Bear & \\
        Sofa & Book& Wine Glass &  Chair  & \\ 
        Cell Phone & Scissors & Cup & & \\
        TV Monitor & Laptop & & & \\
        \bottomrule
    \end{tabular}
\end{table}

\subsection{TUM RGB-D Experiment Setup} \label{subsec:tum}
We used the same 4 sequences from the TUM RGB-D dataset that were used by \cite{Nicholson2018} (\emph{fr1/desk}, \emph{fr1/desk2}, \emph{fr2/desk} and \emph{fr3/long\_office\_household}). We compared the Absolute Trajectory Error (ATE) between standalone QuadricSLAM, QuadricSLAM with both orientation factors and ORB-SLAM2. 

For our setup, visual odometry was sourced from ORB-SLAM2 \cite{Mur-Artal2016} with loop closures disabled. YOLOv3 \cite{YOLO}\cite{Redmon2018} was used for object detection with pretrained weights and associations between individual detections and objects were provided by a set of manual annotations. In classifying the dual quadrics, we combined the classification of individual detections by averaging the full set of detection scores belonging to each object. We then assigned the most likely class as the object's final classification.

Performance loss in original version of QuadricSLAM was caused by poor bounding box estimates and object occlusions. Therefore we also rejected objects with bounding box width or height standard deviation of 50 pixels and greater in order to reduce the number of quadric estimates that require significant adjustments to their orientation. 

In these sequences, quadrics are estimated relative to the camera coordinate frame as opposed to the global coordinate frame. Given that our geometric factors are orienting quadrics with respect to the global frame we need to transform quadric estimates from the camera frame to the global frame. We do this by using the transformation matrix of the first camera pose and its ground truth to determine the pose of quadric estimates relative to the global frame. Ideally we would like to incorporate IMU measurements into QuadricSLAM in order to determine the transformation between the camera frame and the global frame for this process.  

\subsection{TUM RGB-D Results} \label{subsec:tumresults}
For this experiment we compared the standard Absolute Trajectory Error (ATE) of standalone QuadricSLAM, QuadricSLAM with both orientation factor and ORB-SLAM2. The results show that using the orientation factors in conjunction with rejecting objects with high bounding box variance was able to improve the trajectory universally in comparison to standalone QuadricSLAM (See Table~\ref{tab:tumresults}). Both \emph{fr2/desk} and \emph{fr3/long\_office\_household} saw significant improvements to their trajectory error, $14.7\%$ and $28.5\%$ respectively. Whereas \emph{fr1/desk} and \emph{fr1/desk2} saw smaller gains of $3.0\%$ and $0.2\%$ respectively. 

\begin{figure*}[t]
    \centering
    \begin{subfigure}[t]{0.5\textwidth}
    \centering
    \includegraphics[height=1.3in]{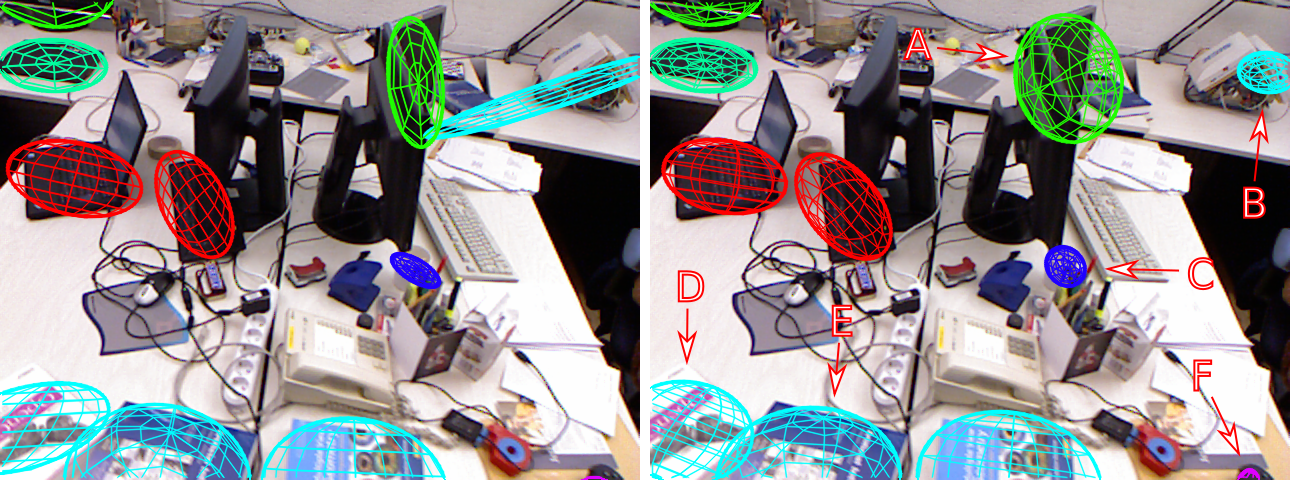}
    \caption{fr1/desk}
    \label{subfig:fr1_desk}
    \end{subfigure}%
    \begin{subfigure}[t]{0.5\textwidth}
    \centering
    \includegraphics[height=1.3in]{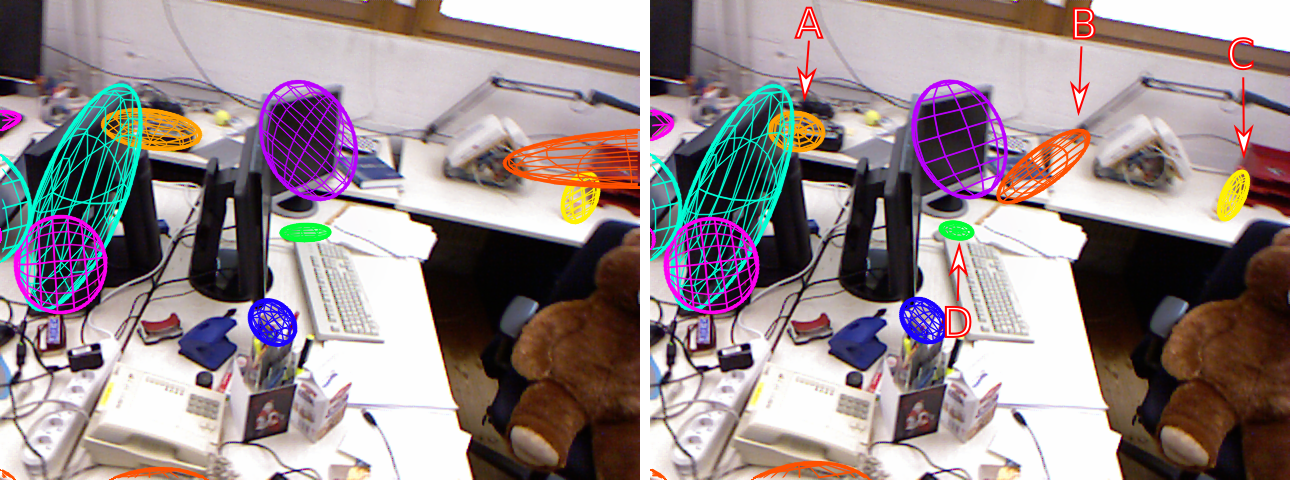}
    \caption{fr1/desk2}
    \label{subfig:fr1_desk2}
    \end{subfigure}
    
    \vspace{0.2cm}
    
    \begin{subfigure}[t]{0.495\textwidth}
    \centering
    \includegraphics[height=1.3in]{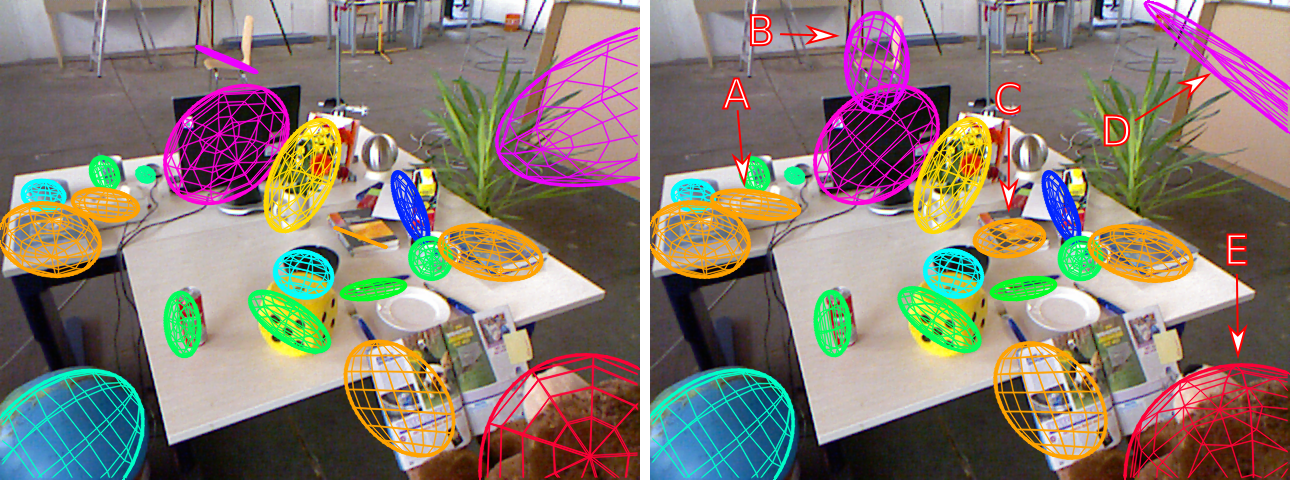}
    \caption{fr2/desk}
    \label{subfig:fr2_desk}
    \end{subfigure}
    \begin{subfigure}[t]{0.495\textwidth}
    \centering
    \includegraphics[height=1.3in]{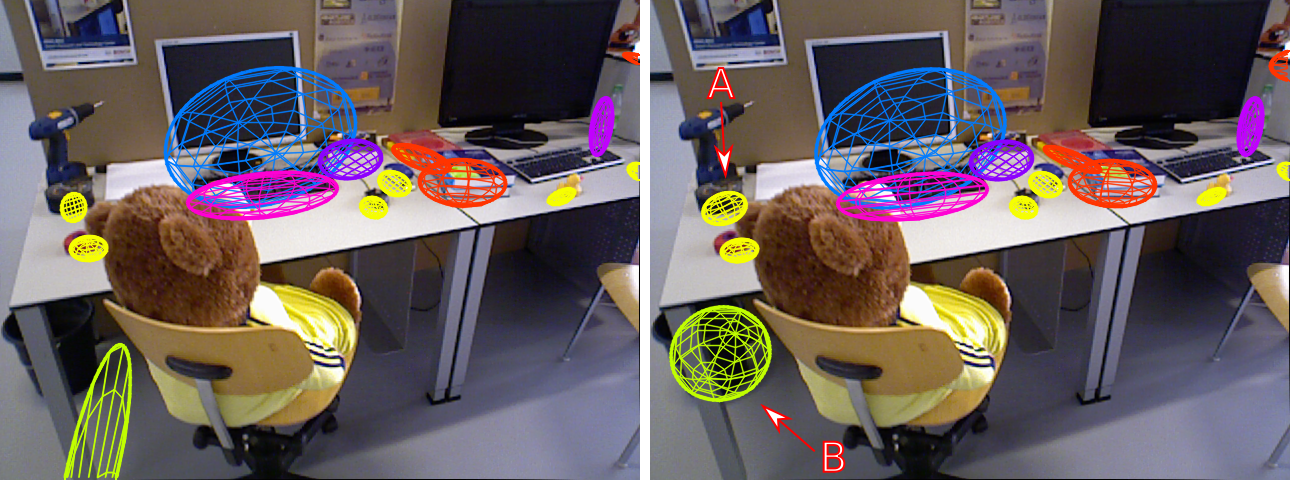}
    \caption{fr3/long\_office\_household}
    \label{subfig:fr3_loh}
    \end{subfigure}
    
    \caption{The estimated quadrics of each sequence, comparing standalone QuadricSLAM (\emph{left}) against QuadricSLAM with both orientation factors (\emph{right}). Occluded objects are not displayed. Notable changes to quadric estimates are indicated with red arrows.}
    \label{fig:tum_comparison}
\end{figure*}

Examining the qualitative result (See Figure~\ref{fig:tum_comparison}), several observations were made. Quadric estimates that were initially unsatisfactory (\ie not fitting around the object it models) were improved with the addition of these factors. Looking at object A in Figure~\ref{subfig:fr1_desk} we can see that the quadric estimation conforms to the monitor's overall shape rather than just the shape of the screen. Object C's estimate of the cup, also in Figure~\ref{subfig:fr1_desk}, shows how the quadric's orientation has changed resulting in a representation that envelops the object and is oriented in the right direction. Other notable examples of this behaviour include object B in Figures~\ref{subfig:fr1_desk2}, \ref{subfig:fr2_desk} and \ref{subfig:fr3_loh}.

\subsubsection{Edge \& Failure Cases}\label{subsubsec:fail}
In some instances quadric estimates did not improve. This is likely due to bounding box and orientation factors imposing on a quadric's shape. However, there were a small number of cases where objects were misclassified and therefore were assign the wrong orientation factor. Object D in Figure~\ref{subfig:fr2_desk} is a large cardboard box that was misclassified as a chair and therefore was assigned a vertical factor instead of a horizontal factor.

For some quadric estimates, it appears that the orientation factors may contribute to a change in translation as well as orientation. This is either beneficial to the estimate (object B in Figure~\ref{subfig:fr3_loh}) or detrimental (object B in Figure~\ref{subfig:fr1_desk} which should be over the book near object A). Overall, the change in translation in this dataset has been beneficial from qualitative perspective. However, because of a lack of ground truth 3D bounding boxes or quadric estimates it's difficult to determine if this behaviour is impacting all quadrics.

\subsubsection{Parameter Selection for Orientation Noise Estimates}
\begin{figure}[t]
    \centering
    \includegraphics[width=\linewidth]{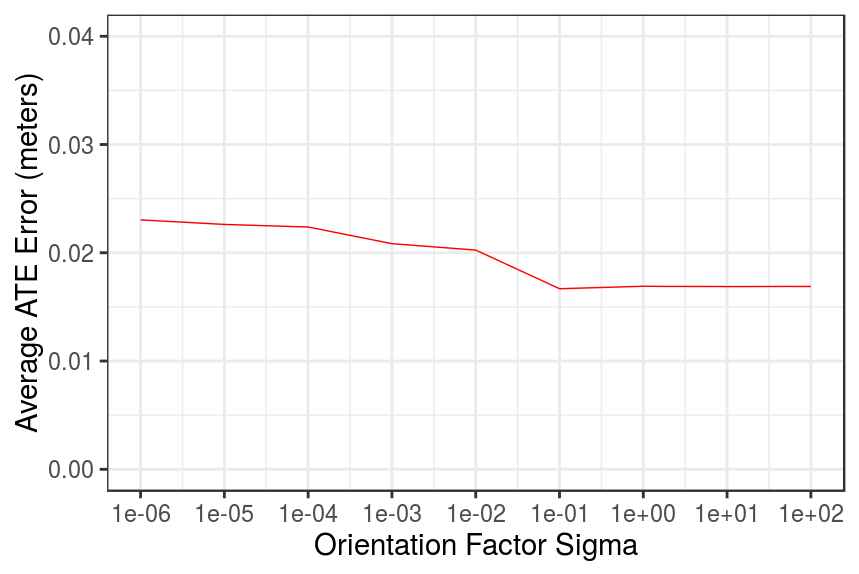}
    \caption{The average ATE error of different orientation factor noise estimates over the four TUM RGB-D sequences.}
    \label{fig:ate_error_tum}
\end{figure}

We tested a range of different values, from $1\mathrm{e}{-6}$ to $1\mathrm{e}{+2}$, for the orientation factor's noise model.
Overall, the choice of value for the orientation factor's noise estimates has no adverse effects on the quality of the trajectory estimation, this is illustrated in Figure~\ref{fig:ate_error_tum}. Moreover, using a value of $1\mathrm{e}{-1}$ or greater has no significant variation in the error. This is beneficial as it shows that the choice of value for this noise estimate is not critical.

\begin{table}[t]
\centering
\caption{A comparison of the average localization errors (in meters) on the TUM RGB-D sequences. fr3/l\_o\_h abbreviates fr3/long\_office\_household. Bold denotes best performance for the sequence.}
\label{tab:tumresults}
\begin{tabular}{@{}lccc@{}}
    \toprule
    Sequence & QuadricSLAM & Both Orientation Factors & ORB-SLAM2 \\ \midrule
    fr1/desk & 0.0167 & 0.0162 & \textbf{0.0159} \\
    fr1/desk2 & 0.0245 & 0.0244 & \textbf{0.0243} \\
    fr2/desk & 0.0124 & 0.0105 & \textbf{0.0087} \\
    fr3/l\_o\_h & 0.0230 & 0.0165 & \textbf{0.0107} \\ \bottomrule
\end{tabular}
\end{table}

\subsection{Simulated Experiment Setup} \label{subsec:sim}
Our simulated environment allowed us to evaluate the landmark quality of the quadrics against the ground truth object positions. This dataset was created using the UnrealCV plugin \cite{Qiu2017} and contains ground truth camera trajectory $\vx_i \in \text{SE(3)}$, 2D object bounding boxes $\vect{b}_{ij}$, and 3D object bounding boxes. The camera was simulated with a focal length of 320.0 and a principal point $(C_x, C_y) = (320.0, 240.0)$ at a resolution of $640 \times 480$. 

As with \cite{Nicholson2018}, we used 50 ground truth trajectories that were recorded over 10 scenes. Each trajectory was injected with noise generated from 5 seeds resulting in a total of 250 trials. The relative motion between trajectory positions was corrupted by zero mean Gaussian noise to induce an error of 5\% for translation and 15\% for rotation. This noise is similar to noise in standard inertial navigation systems. Object detections were simulated by adding additional variance of 4 pixels to the ground truth bounding box detections which was used in the original QuadricSLAM experiments. 

We evaluated the estimated quadric parameters against the ground truth 3D bounding boxes. We used the three landmark metrics from \cite{Nicholson2018} for this experiment to measure the position, shape and quality of the quadric estimates. Specifically, the metrics we used:

\begin{itemize}
\item \textbf{Landmark position} is evaluated by calculating the root mean squared error (RMSE) between the estimated quadric's centroid and the ground truth 3D bounding box centroid. 
\item \textbf{Landmark shape} is evaluated by using the Jaccard distance ($1 - \text{Intersection over Union}$) between estimated quadric's 3D axis aligned bounding box and the ground truth bounding box after centering both boxes at the origin. 
\item \textbf{Landmark quality} is evaluated by using the standard Jaccard distance between the estimated and true bounding boxes. This metric takes into account the position, shape and orientation of the landmark.
\end{itemize}

\subsection{Simulated Results} \label{subsec:simresults}
We compared the three landmark metrics of standalone QuadricSLAM and QuadricSLAM with both orientation factors. The results showed that overall that there was no improvement to the landmark quality. The inclusion of the orientation factors did improve the average shape of the landmarks by $0.2\%$. However, landmark position and quality saw a decrease of $1.2\%$ and $0.3\%$ respectively. This suggests that some quadric estimates are translated when imposed with the orientation. This is consistent with the observations made in Section~\ref{subsec:tumresults}. Furthermore, the landmark quality metric is influenced by landmark position and shape which explains the increase in the quality error. 

\begin{table}[t]
    \centering
    \caption{A comparison of the average errors for landmark estimates in the simulated environment. Bold denotes best performance for the metric.}
    \label{tab:simresults}
    \begin{tabular}{@{}lccc@{}}
        \toprule
         & Position (m) & Shape (\%) & Quality (\%)  \\ \midrule
        QuadricSLAM & \textbf{0.1714} & 0.4450 & \textbf{0.5886} \\ 
        Both Orientation Factors & 0.1739 & \textbf{0.4445} & 0.5909 \\\bottomrule
    \end{tabular}
\end{table}

\section{Conclusion \& Future Work} \label{sec:conclusion}

We developed a geometric factor that exploits prior knowledge of the world to constrain the orientation of quadric estimates in QuadricSLAM. 
Such geometric constraints for object-oriented SLAM support the accurate representation of objects in the resulting map -- essential if object-oriented maps are going to be used for robotic scene understanding.

Our proposed factor aligns the major axis of the quadrics to be either perpendicular to or aligned with the direction of gravity. We used the semantics of the landmark objects to better inform the geometry of the quadrics by applying the appropriate orientation factor based on object classification. 

Our paper demonstrated the benefits of geometric factors both quantitatively and qualitatively for object-oriented maps. We performed an extensive evaluation of the trajectory and landmark quality using both real-world and synthetic datasets. In using these factors, we were able the improve the trajectory as well as the quality of the landmark shape to standalone QuadricSLAM. 

Future work will investigate the applicability of this method to different environments, other geometric factors that can be used to improve landmark quality, identifying the appropriate orientation of an unknown object in open set conditions, methods that can learn the correct orientation of objects, and the utility of object-oriented maps for tasks like manipulation or navigation.

\bibliographystyle{IEEEtran}
\bibliography{ref}

\end{document}